\documentclass[conference]{IEEEtran}
\ifdefined\pdfinclusioncopyfonts
\fi
\ifdefined\pdfminorversion
\fi
\IEEEoverridecommandlockouts 
\usepackage[T1]{fontenc}
\usepackage{ae,aecompl}
\usepackage[english]{babel}
\usepackage[utf8]{inputenc}
\usepackage{times} 
\usepackage{cite}

\usepackage{booktabs}

\usepackage[bookmarks=false]{hyperref}
\usepackage[nolist]{acronym}

\usepackage{multirow}

\usepackage{subcaption}

\usepackage{siunitx}
\usepackage{graphicx} 
\usepackage{epsfig} 
\usepackage{pgfplots}
\pgfplotsset{compat=1.18}
\usepgfplotslibrary{statistics}
\usepgfplotslibrary{groupplots}

\usepackage{float}

\usepackage[export]{adjustbox}

\usepackage{xcolor}
\definecolor{myYellow}{rgb}{0.93,0.69,0.13}
\definecolor{myPurple}{rgb}{0.49,0.18,0.56}
\definecolor{myGreen}{rgb}{0.26 0.72 0.54}
\definecolor{darkgreen}{rgb}{0.272, 0.50, 0.376}
\definecolor{lightgreen}{rgb}{0.585, 0.82, 0.647}

\colorlet{mydarkblue}{blue!30!black}

\usepackage{mathtools}
\usepackage{nicefrac}
\usepackage{amsmath}
\usepackage{amsthm}
\usepackage{amssymb}
\usepackage{bm} 
\usepackage{scalerel} 

\DeclareMathAlphabet{\pazocal}{OMS}{zplm}{m}{n}

\usepackage{etoolbox}
\makeatletter%
\AfterPreamble{%
	\usepackage{hyperref}%
	\let\oldhypertarget\hypertarget%
	\renewcommand{\hypertarget}[2]{%
		\oldhypertarget{#1}{#2}%
		\protected@write\@mainaux{}{%
			\string\expandafter\string\gdef%
			\string\csname\string\detokenize{#1}\string\endcsname{#2}%
		}%
	}%
	\newcommand{\myhyperlink}[1]{%
		\hyperlink{#1}{\csname #1\endcsname}%
	}%
    \renewcommand{\hypertarget}[2]{%
    \oldhypertarget{#1}{#2}%
    \protected@write\@mainaux{}{%
        \string\expandafter\string\gdef%
        \string\csname\string\detokenize{#1}\string\endcsname{\arabic{Theorem}}%
    }%
}%
}
\makeatother%

\newcounter{Remark}
\newcommand{\displayRemarks}[2][]{%
	\stepcounter{Remark}%
	\textbf{Remark}~\hypertarget{#1}{\textbf{\theRemark}}{~(#2)}%
}

\newcounter{Problem}
\newcounter{Theorem}
\makeatletter
\newcommand{\displayTheorem}[2][]{%
    \stepcounter{Theorem}%
    \hypertarget{#1}{}
    \textbf{Theorem~\theTheorem}~(#2)%
}
\makeatother

\newcommand{\refTheorem}[1]{%
	Theorem~\myhyperlink{#1}%
}

\newcounter{Lemma}
\makeatletter
\newcommand{\displayLemma}[2][]{%
    \stepcounter{Lemma}%
    \hypertarget{#1}{}
    \textbf{Lemma~\theLemma}~(#2)%
}
\makeatother

\usepackage{algorithm}
\usepackage[noend]{algpseudocode}

\makeatletter
\def\BState{\State\hskip-\ALG@thistlm}
\makeatother

\usepackage{tikz}
\pgfdeclarelayer{foreground}
\pgfsetlayers{background, main, foreground}
\usetikzlibrary{quotes, angles, backgrounds, arrows, automata, shapes, positioning, calc, through, spy, decorations.pathreplacing, decorations.markings, arrows.meta, automata, petri, shapes.multipart}

\tikzset{
    imglabel/.style={
      rectangle,
      inner sep=2pt,
      text=black,
      minimum height=1em,
      text centered,
      fill=white,
      fill opacity=1.0,
      text opacity=1,
      anchor=south west,
    },
  }
\tikzset{
	state/.style={
		rectangle,
		draw=black, very thick,
		minimum height=1.0em,
		text centered,
	},
}
\tikzset{
  on each segment/.style={
    decorate,
    decoration={
      show path construction,
      moveto code={},
      lineto code={
        \path [#1]
        (\tikzinputsegmentfirst) -- (\tikzinputsegmentlast);
      },
      curveto code={
        \path [#1] (\tikzinputsegmentfirst)
        .. controls
        (\tikzinputsegmentsupporta) and (\tikzinputsegmentsupportb)
        ..
        (\tikzinputsegmentlast);
      },
      closepath code={
        \path [#1]
        (\tikzinputsegmentfirst) -- (\tikzinputsegmentlast);
      },
    },
  },
  mid arrow/.style={postaction={decorate,decoration={
        markings,
        mark=at position .5 with {\arrow[#1]{stealth}}
      }}},
}

\tikzset{
  half circle/.style={
      semicircle,
      shape border rotate=180,
      anchor=chord center,
      minimum size=5mm
      }
}

\def\BibTeX{{\rm B\kern-.05em{\sc i\kern-.025em b}\kern-.08em
    T\kern-.1667em\lower.7ex\hbox{E}\kern-.125emX}}  

\newcommand\copyrightnotice{%
	\begin{tikzpicture}[remember picture,overlay]
	\node[anchor=south,yshift=26.0cm] at (current page.south) 
	{\color{red}\fbox{\parbox{\dimexpr\textwidth-\fboxsep-\fboxrule\relax}{\copyrighttext}}};
	\end{tikzpicture}%
}    

\tikzset{
	on each segment/.style={
		decorate,
		decoration={
			show path construction,
			moveto code={},
			lineto code={
				\path [#1]
				(\tikzinputsegmentfirst) -- (\tikzinputsegmentlast);
			},
			curveto code={
				\path [#1] (\tikzinputsegmentfirst)
				.. controls
				(\tikzinputsegmentsupporta) and (\tikzinputsegmentsupportb)
				..
				(\tikzinputsegmentlast);
			},
			closepath code={
				\path [#1]
				(\tikzinputsegmentfirst) -- (\tikzinputsegmentlast);
			},
		},
	},
	mid arrow/.style={postaction={decorate,decoration={
				markings,
				mark=at position .5 with {\arrow[#1]{stealth}}
	}}},
}

\tikzset{
	half circle/.style={
		semicircle,
		shape border rotate=180,
		anchor=chord center,
		minimum size=5mm
	}
}

\newcommand\copyrighttext{%
	\small \begin{center}\vspace{-0.5em} \color{red} \textcopyright\,2026 IEEE. Accepted for presentation to the ``2026 IEEE Global Communications Conference (GLOBECOM)", 7–11 December 2026, Macau S.A.R., China. Personal use of this material is permitted. Permission from IEEE must be obtained for all other uses, in any current or future media, including reprinting/republishing this material for advertising or promotional purposes, creating new collective works, for resale or redistribution to servers or lists, or reuse of any copyrighted component of this work in other works.\vspace{-0.75em} \end{center}
}

\begin{document}

\title{\copyrightnotice \LARGE \bf 3D Euler-Angle Orientation Control for Two-Ray Fading Mitigation in Maritime Air-to-Sea Communications
}

\author{Mohammed Bajja$^{1\star}$, Abdoul Karim A. H. Saliah$^{1\star}$, Hajar El Hammouti$^{1}$, \\Daniel Bonilla Licea$^{1}$, and Giuseppe Silano$^{2}$
 \thanks{$^1$Authors are with the Mohammed VI Polytechnic University, Ben Guerir, Morocco (emails: {\tt\footnotesize \{mohammed.bajja, abdoul.saliah, hajar.elhammouti, daniel.bonilla\}@um6p.ma}).}
  \thanks{$^2$Authors are with Ricerca sul Sistema Energetico S.p.A., Milan, Italy, and Czech Technical University in Prague, Prague, Czech Republic (email: {\tt\footnotesize giuseppe.silano@fel.cvut.cz}).}
  \thanks{Partially funded by the Italian Electrical System research fund (decree n.~388, Nov.~6, 2024), the EU grant no.~DCI-PANAF/2020/420-028 (ARISE), the CTU grant no.~SGS26/077/OHK3/1T/13, and the GA\v{C}R project no.~26-22606S.
  $^\star$Equal contribution.} 
}

\maketitle
\thispagestyle{empty} 
\pagestyle{empty} 


\begin{acronym}
    \acro{A2S}[A2S]{Air-to-Sea}
    \acro{BS}[BS]{Base Station}
    \acro{CoM}[CoM]{Center of Mass}
    \acro{GTMR}[GTMR]{Generically Tilted Multi-Rotor}
    \acro{LoS}[LoS]{Line-of-Sight}
    \acro{NMPC}[NMPC]{Nonlinear Model Predictive Control}
    \acro{SNR}[SNR]{Signal-to-Noise-Ratio}
    \acro{UAS}[UAS]{Unmanned Aircraft System}
    \acro{UAV}[UAV]{Unmanned Aerial Vehicle}
    \acro{wrt}[w.r.t.]{with respect to}
\end{acronym}



\begin{abstract}

    Maritime \acl{A2S} links are dominated by a line-of-sight ray and a sea-surface reflected ray whose destructive combination produces deep fades. Existing mitigation strategies optimize \acl{UAV} position or trajectory but leave attitude unexploited. This paper treats the full three-dimensional attitude as a physical-layer control variable that shapes the two-ray interference through antenna phase-center displacement. Under small-angle and far-field assumptions, the constructive-interference condition reduces to an affine constraint in the Euler angles and admits a closed-form family 
    of minimum-norm attitude candidates. A differentiable soft-minimum rule yields a smooth reference tracked by a constrained \acl{NMPC} controller on a fully-actuated tilting multirotor. The proposed scheme increases cumulative throughput by 11.4\% over a pitch-only benchmark and 22.2\% over a zero-orientation baseline, while preserving trajectory tracking and respecting actuator limits.

\end{abstract}



\section{Introduction}
\label{sec:introduction}

Reliable wireless connectivity over open sea is a critical enabler for offshore monitoring, search-and-rescue, infrastructure inspection, and beyond-visual-line-of-sight 
logistics \cite{Alqurashi2023MaritimeSurvey, Nomikos2023UAVMaritimeSurvey}. 
\acfp{UAV} are well suited to such missions as rapidly deployable aerial nodes \cite{MAHBUB2025100977}, and a large body of work has addressed \ac{UAV}-enabled communications through channel characterization, deployment, and trajectory design \cite{10616106, MAHBUB2025100977}. Maritime links, however, exhibit propagation mechanisms 
markedly different from terrestrial settings: the \ac{LoS} component is typically strong, 
and the dominant additional component is the sea-surface reflection, yielding the classical \textit{two-ray interference pattern}
\footnote{The two-ray model superposes a \ac{LoS} ray and a sea-surface reflection whose relative phase produces deep fades.} and its 
curved-earth refinements \cite{Matolak2017UASOverWaterPartI, 11267247, Liu2021NonStat6GUAVMaritime}. 
The resulting deep fades can severely degrade link quality even when average path loss is moderate \cite{11267247}.

A defining feature of two-ray \acf{A2S} channels is their extreme sensitivity to geometry: small changes in platform position or orientation produce non-negligible variations in the 
relative phase between the \ac{LoS} and reflected paths, and therefore large instantaneous \ac{SNR} fluctuations \cite{Wang2022UAVJitter, Yang2025U2SSeaClutter}. \emph{Instantaneous} link performance can thus deviate substantially from average path-loss predictions, and pose-induced perturbations become part of the effective channel dynamics. Beyond planning \emph{where} the \ac{UAV} should fly, this raises the natural question of whether \emph{how} it is oriented can be exploited to actively regulate the dominant interference mechanism. 

Most existing designs mitigate two-ray fading indirectly by optimizing position, altitude, and trajectory \cite{ MAHBUB2025100977}, while measurement-driven models describe deep fades without exploiting them \cite{Matolak2017UASOverWaterPartI, 11267247}. A critical aspect remains under-exploited: \emph{treating \ac{UAV} attitude as a controllable degree of freedom that intentionally shapes the instantaneous two-ray interference condition}. 
Our work takes the same physical insight as \cite{Wang2022UAVJitter}---that attitude variations perturb the antenna phase-center location and hence the two-ray phase difference---but inverts its use: 
whereas \cite{Wang2022UAVJitter} characterizes the attitude jitter as a random process, we treat the attitude as a \emph{control input},
computing a deterministic 3D Euler-angle reference that proactively steers the two-ray phase toward 
constructive interference. Unlike pose optimization in adversarial settings 
\cite{BonillaLicea2024PoseICASSP, BonillaLicea2024CoopJamming, Silano2026CommNMPCJamming}, our objective is the deterministic phase relationship of the two-ray geometry, which yields a closed-form expression for the optimal attitude.

This paper extends the emerging viewpoint of \emph{communications-aware robotics}, in which motion and control are co-designed with communication objectives \cite{BonillaIEEEProc, BonillaLicea2025IEEEComMag}. In long-range maritime regimes, centimeter-scale pose-induced antenna displacements produce appreciable phase shifts at typical carrier wavelengths (e.g., $\lambda \approx 10$\,cm at $3$\,GHz), moving the operating point across constructive/destructive interference fringes. We design an attitude strategy that (i) yields an explicit 3D Euler-angle reference maximizing instantaneous received \ac{SNR} under small-angle and far-field assumptions, and (ii) is tracked by a constrained controller that preserves mission-level motion objectives. Fully-actuated multirotor platforms \cite{Hamandi2021MRAV, Ryll2019Interaction} are key enablers, as they decouple attitude regulation from translational authority. We adopt \acf{NMPC} \cite{Mayne2000MPCStability, SilanoSMC2025} with established real-time numerical schemes \cite{SilanoSMC2025, Silano2026CommNMPCJamming} to handle dynamics, constraints, and reference tracking in a unified problem.

The main contributions of this paper are as follows:
\begin{itemize}
    \item \textbf{3D orientation-aware \ac{A2S} two-ray model:} 
    We extend the two-ray model to capture how roll, pitch, and yaw displace the antenna phase center and shift the instantaneous interference condition, exposing attitude as a communication control variable.

    \item \textbf{Closed-form Euler-angle optimization:} Under small-angle and far-field 
    assumptions, the constructive-interference condition reduces to an affine constraint in the Euler angles, governed by a geometry-dependent phase-sensitivity vector that yields closed-form minimum-norm attitude candidates.

    \item \textbf{Smooth 3D attitude reference law:} A differentiable soft-minimum rule selects the smallest feasible candidate among constructive-interference solutions, avoiding discontinuous switching while remaining compatible with real-time control.

    \item \textbf{Communications-aware \ac{NMPC} tracking:} The resulting reference is embedded in a constrained \ac{NMPC} controller for a fully-actuated tilting 
    multirotor, enabling simultaneous trajectory tracking and link-quality enhancement without modifying the position reference.
\end{itemize}



\section{System Model}
\label{sec:systemModel}

\vspace{4pt}
\begin{figure}[tb]
\vspace{6pt}
    \centering
    \includegraphics[width=0.86\columnwidth]{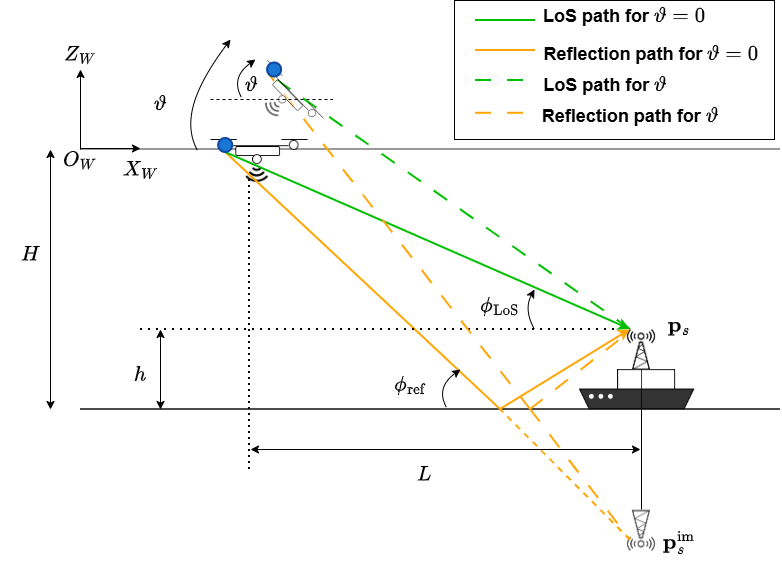}
    \vspace*{-0.65em}
    \caption{2D vertical cross-section with pitch-only motion; The antenna center phase is depicted as a blue marker.
    }
    \label{fig:System_Mod}
\end{figure}

We consider a \ac{UAV} communicating with a ship through an \ac{A2S} two-ray channel (see Fig. \ref{fig:System_Mod}). We adopt an inertial world frame $\mathcal{F}_W$ whose $z$-axis is vertical and whose origin is set at the UAV flight altitude, so that points below it have negative $z$-coordinates; in particular, the sea surface is at $z=-20~\mathrm{m}$.



\subsection{Two-ray channel model}
\label{Classical_2ray_Sys_Mod}

We adopt the standard two-ray \ac{A2S} abstraction~\cite{Wang2022UAVJitter}: (i) the received signal is the superposition of a direct \ac{LoS} ray and a single sea-surface specular reflection; (ii) the reflection is modeled by a constant magnitude factor $\gamma\in(0,1)$ and an additional $\pi$ phase shift; (iii) sea-state-induced diffuse scattering, shadowing by waves and ship superstructure, polarization mismatch, and Doppler are neglected. Both ends are equipped with isotropic antennas, assumed for analytical tractability. These simplifications isolate the geometric mechanism that governs deep fades and admit closed-form orientation laws.

Let $H$ be the altitude of the \ac{UAV} antenna phase center, $h$ that of the ship antenna, and $L$ the horizontal separation between them. The \ac{LoS} and specular path lengths are $d_{\rm{LoS}}=\sqrt{L^2+(H-h)^2}$ and $d_{\rm{ref}}=\sqrt{L^2+(H+h)^2}$.
The reflected-minus-direct phase difference is $\Delta\Phi = \tfrac{2\pi}{\lambda}\bigl(d_{\rm{ref}}-d_{\rm{LoS}}\bigr)$,
where $\lambda$ is the carrier wavelength. The two-ray baseband channel coefficient is~\cite{Wang2022UAVJitter}
\begin{equation}\label{eq:g_static}
    g = \tfrac{\lambda}{4\pi}\bigl(d_{\rm{LoS}}^{-1} + \gamma\,d_{\rm{ref}}^{-1}\,e^{-j(\Delta\Phi+\pi)}\bigr),
\end{equation}
and the received power is $P_r = P_t|g|^2$, with $P_t$ the transmit power. Deep fades arise when the two contributions combine destructively, i.e., when $\Delta\Phi \approx 2\pi m$ for some $m\in\mathbb{Z}$, in which case the $\pi$ reflection phase causes near-cancellation despite both rays being individually strong.



\subsection{Orientation-dependent channel}
\label{Orientation_2ray_Sys_Mod}

Equation~\eqref{eq:g_static} treats the \ac{UAV} as a point. We now make the dependence on attitude explicit. Let $\mathbf{p}_U\in\mathbb{R}^3$ be the position of the \ac{UAV} \ac{CoM} in $\mathcal{F}_W$, and let $\bm{\eta}=[\phi,\vartheta,\psi]^\top$ collect the roll, pitch, and yaw angles defining the body frame $\mathcal{F}_B$. The antenna phase center is mounted at a fixed offset $\mathbf{r}_a\in\mathbb{R}^3$ in $\mathcal{F}_B$, so its world-frame position is
\begin{equation}
    \mathbf{p}_a(\bm{\eta})
    =
    \mathbf{p}_U+\mathbf{R}(\bm{\eta})\mathbf{r}_a,
    \label{eq:antenna_phase_center_3d}
\end{equation}
where $\mathbf{R}(\bm{\eta})\in\mathrm{SO}(3)$ is the body-to-world rotation matrix. 

The ship antenna is located at $\mathbf{p}_s=[x_s,y_s,h]^\top$. Treating the sea as a horizontal infinite reflector, the image-source method places the virtual receiver at $\mathbf{p}_s^{\rm im}=[x_s,y_s,-h]^\top$, so that the orientation-dependent path lengths are
\begin{align}
    d_{\rm{LoS}}(\bm{\eta}) = \|\mathbf{p}_s - \mathbf{p}_a(\bm{\eta})\|,\quad 
    d_{\rm{ref}}(\bm{\eta}) = \|\mathbf{p}_s^{\rm im} - \mathbf{p}_a(\bm{\eta})\|.
\end{align}
The phase difference and channel coefficient inherit the orientation dependence:
\begin{equation}\label{eq:DeltaPhi_eta}
    \Delta\Phi(\bm{\eta})
    =
     \tfrac{2\pi}{\lambda}\bigl(d_{\rm{ref}}(\bm{\eta}) - d_{\rm{LoS}}(\bm{\eta})\bigr),
\end{equation}
\begin{equation}\label{eq:g_eta}
    g(\bm{\eta}) = \tfrac{\lambda}{4\pi}\Bigl(d_{\rm{LoS}}^{-1}(\bm{\eta}) + \gamma\,d_{\rm{ref}}^{-1}(\bm{\eta})\,e^{-j(\Delta\Phi(\bm{\eta})+\pi)}\Bigr).
\end{equation}
At $\bm{\eta}=\mathbf{0}$, eqs~\eqref{eq:DeltaPhi_eta}--\eqref{eq:g_eta} recover the static model of Sec.~\ref{Classical_2ray_Sys_Mod}. The attitude thus enters the channel through both the path-length and the phase term; in the far-field maritime regime considered here, the phase term dominates (see Sec.~\ref{sec:theta_opt}).



Since $P_r \propto |g(\bm{\eta})|^2$, maximizing the received power amounts to selecting the attitude
\begin{equation}\label{eq:opt_exact_eta}
    \bm{\eta}^\star \in 
    \arg\max_{\bm{\eta} \in \mathcal{E}} 
    \,\bigl|g(\bm{\eta})\bigr|^2,
\end{equation}
where $\mathcal{E}\subset\mathbb{R}^3$ is the admissible Euler-angle set imposed by mechanical limits, flight safety, and the small-angle operating regime adopted below. Problem~\eqref{eq:opt_exact_eta} is non-convex due to the trigonometric dependence of $\mathbf{R}(\bm{\eta})$, $d_{\rm{LoS}}$, and $d_{\rm{ref}}$ on $\bm{\eta}$. Section~\ref{sec:theta_opt} derives a tractable closed-form approximation.



\section{Closed-Form 3D Orientation Optimization}
\label{sec:theta_opt}

The exact problem~\eqref{eq:opt_exact_eta} is non-convex. We derive a tractable closed-form approximation by exploiting two regularities of long-range maritime \ac{A2S} links: the attitude corrections of interest are small, and the link distance is large compared to the antenna offset. Section~\ref{Assumptions_And_Simplification} states these assumptions and linearizes the two-ray phase. Section~\ref{Closed_Form_Solution} uses the linearized model to obtain a closed-form family of minimum-norm attitude candidates that enforce constructive interference at leading order.



\subsection{Far-field small-angle regime}
\label{Assumptions_And_Simplification}

\textbf{(A1) Small-angle regime:} The required attitude corrections satisfy $\|\bm{\eta}\| \ll 1$ rad, so that the rotation matrix admits the first-order expansion
\begin{equation}\label{eq:small_angle_rotation}
    \mathbf{R}(\bm{\eta})\,\mathbf{r}_a 
    = 
    \mathbf{r}_a + \mathbf{J}_a\,\bm{\eta} 
    + \mathcal{O}(\|\bm{\eta}\|^2),
\end{equation}
where $\mathbf{J}_a = -[\mathbf{r}_a]_\times \in \mathbb{R}^{3\times 3}$ is the negative skew-symmetric matrix associated with $\mathbf{r}_a$, with columns $-\mathbf{e}_i \times \mathbf{r}_a$ for $i=1,2,3$.

\textbf{(A2) Far-field regime:} The link distance is large compared to the antenna offset, $\|\mathbf{p}_s - \mathbf{p}_{a,0}\| \gg \|\mathbf{r}_a\|$, where $\mathbf{p}_{a,0} = \mathbf{p}_a(\mathbf{0})$. Under this assumption, the rays in a neighborhood of the nominal antenna position can be treated as locally parallel, with directions
\begin{equation}\label{eq:unit_vectors}
    \mathbf{u}_{\rm{LoS}} 
    = 
    \frac{\mathbf{p}_s - \mathbf{p}_{a,0}}{\|\mathbf{p}_s - \mathbf{p}_{a,0}\|},
    \quad
    \mathbf{u}_{\rm{ref}} 
    = 
    \frac{\mathbf{p}_s^{\rm im} - \mathbf{p}_{a,0}}{\|\mathbf{p}_s^{\rm im} - \mathbf{p}_{a,0}\|}.
\end{equation}

Under \textbf{(A1)}--\textbf{(A2)}, the path-length variations \acl{wrt} the nominal pose are well approximated by projecting the antenna displacement~\eqref{eq:small_angle_rotation} onto these directions, yielding an affine model of the two-ray phase derived in the next subsection.



\subsection{Closed-form minimum-norm candidates}
\label{Closed_Form_Solution}

Expanding the squared magnitude of~\eqref{eq:g_eta} gives
\begin{equation}\label{eq:power_expand}
\resizebox{0.90\hsize}{!}{$
    |g(\bm{\eta})|^2 
    = 
    \left(\dfrac{\lambda}{4\pi}\right)^{\!2}
    \Bigg(
    \dfrac{1}{d_{\rm{LoS}}^2(\bm{\eta})}
    + \dfrac{\gamma^2}{d_{\rm{ref}}^2(\bm{\eta})} 
    - \dfrac{2\gamma\cos\bigl(\Delta\Phi(\bm{\eta})\bigr)}{d_{\rm{LoS}}(\bm{\eta})\,d_{\rm{ref}}(\bm{\eta})}
    \Bigg).
$}
\end{equation}
Under \textbf{(A1)}--\textbf{(A2)}, the path-length factors in~\eqref{eq:power_expand} vary at $\mathcal{O}(\|\bm{\eta}\|)$ relative to their nominal values, while the cosine term varies at $\mathcal{O}(1)$ as $\Delta\Phi$ traverses one fringe. The dominant orientation dependence is therefore carried by the phase term, and maximizing $|g(\bm{\eta})|^2$ at leading order reduces to minimizing $\cos(\Delta\Phi(\bm{\eta}))$. The minimum is attained when the two contributions to~\eqref{eq:g_eta} combine constructively, despite the $\pi$ reflection phase shift, i.e., when $\Delta\Phi(\bm{\eta}) \in \{(2m+1)\pi : m \in \mathbb{Z}\}$.

The following theorem makes this precise and yields a closed-form family of minimum-norm attitude candidates.

\displayTheorem[thm:eta-candidates]{Closed-form 3D constructive-interference attitudes}
\textit{Under assumptions \textbf{(A1)}--\textbf{(A2)}, the two-ray phase admits the affine approximation
\begin{equation}\label{eq:phase_affine}
    \Delta\Phi(\bm{\eta}) 
    = 
    \Delta\Phi_0 + \mathbf{k}^\top \bm{\eta} + \mathcal{O}(\|\bm{\eta}\|^2),
\end{equation}
with $\Delta\Phi_0 = \Delta\Phi(\mathbf{0})$ and \emph{phase-sensitivity vector}
\begin{equation}\label{eq:k_vector}
    \mathbf{k} 
    = 
    \frac{2\pi}{\lambda}\,\mathbf{J}_a^\top
    \bigl(\mathbf{u}_{\rm{LoS}} - \mathbf{u}_{\rm{ref}}\bigr).
\end{equation}
Assume $\mathbf{k} \neq \mathbf{0}$. Then, at leading order in $\|\bm{\eta}\|$, the attitudes that maximize $|g(\bm{\eta})|^2$ form the family of hyperplanes 
\begin{equation}\label{eq:hyperplane_eta}
    \mathbf{k}^\top \bm{\eta} 
    = 
    C_m, 
    \quad 
    C_m = (2m+1)\pi - \Delta\Phi_0,
    \quad
    m \in \mathbb{Z},
\end{equation}
and the minimum-norm attitude on each hyperplane is
\begin{equation}\label{eq:eta_candidates}
    \bm{\eta}_m 
    = 
    \frac{C_m}{\|\mathbf{k}\|^2}\,\mathbf{k}.
\end{equation}
If $\bm{\eta}_m \in \mathcal{E}$, then $\bm{\eta}_m$ is the feasible attitude of smallest Euclidean norm enforcing constructive interference at leading order.
}

\textit{Proof.} See Appendix~\ref{Appendix_Proof_Theorem}.\hfill$\blacksquare$

The vector $\mathbf{k}$ quantifies the first-order sensitivity of the two-ray phase to roll, pitch, and yaw. Its direction defines the locally most efficient axis of attitude correction, and its magnitude scales inversely with the angular effort required to traverse one interference fringe. The candidates~\eqref{eq:eta_candidates} are aligned with $\mathbf{k}$ and indexed by the integer $m$, which selects the fringe to be matched.

\displayRemarks[rem:degenerate]{Degenerate case}
The condition $\mathbf{k} = \mathbf{0}$ occurs when $\mathbf{J}_a^\top(\mathbf{u}_{\rm{LoS}} - \mathbf{u}_{\rm{ref}}) = \mathbf{0}$, i.e., when the difference of unit directions toward the receiver and its image lies in the null space of $\mathbf{J}_a^\top$. Geometrically, this means no infinitesimal rotation displaces the antenna along the direction that distinguishes the \ac{LoS} and reflected rays, so the two-ray phase is not first-order controllable through attitude. The controller defaults to $\bm{\eta}^\star = \mathbf{0}$, and link improvement must be sought through translational motion.



\subsection{Smooth attitude reference law}
\label{sec:Smooth_Control_Law}

\refTheorem{thm:eta-candidates} yields a countable family of constructive-interference candidates $\{\bm{\eta}_m\}_{m \in \mathbb{Z}}$. Two practical issues arise. First, the small-angle assumption \textbf{(A1)} restricts attention to candidates with sufficiently small norm, motivating a truncation to a finite index set $\mathcal{M} \subset \mathbb{Z}$ chosen so that the induced candidate set $\{\bm{\eta}_m\}_{m \in \mathcal{M}}$ spans the admissible range. Second, the candidates compatible with the mechanical and operational box $\mathcal{E}$ form the feasible subset $\mathcal{M}_f = \bigl\{ m \in \mathcal{M} : \bm{\eta}_m \in \mathcal{E} \bigr\}.$

The natural reference is the feasible candidate of smallest norm, $\bm{\eta}_{m^\star}$ with $m^\star = \arg\min_{m \in \mathcal{M}_f} \|\bm{\eta}_m\|$, which preserves consistency with \textbf{(A1)} and minimizes deviation from the trim attitude. Selecting $\bm{\eta}_{m^\star}$ via $\arg\min$, however, is discontinuous in the candidates: as the link geometry evolves along the mission, two candidates may swap rank, causing the reference to jump and inducing transients that the tracking controller must absorb. We therefore replace $\arg\min$ by a differentiable approximation.

\displayLemma[lem:soft-theta]{Soft minimum-norm selection}\textit{
Let $\{\bm{\eta}_m\}_{m \in \mathcal{M}_f}$ be a finite, nonempty set of feasible candidates with a unique minimum-norm element $\bm{\eta}_{m^\star}$. For $\tau > 0$, define
\begin{equation}\label{eq:theta_soft_min}
    \bm{\eta}_\tau^\star 
    = 
    \frac{\sum_{m \in \mathcal{M}_f} \bm{\eta}_m\,e^{-\tau\|\bm{\eta}_m\|^2}}
         {\sum_{m \in \mathcal{M}_f} e^{-\tau\|\bm{\eta}_m\|^2}}.
\end{equation}
Then $\bm{\eta}_\tau^\star$ is a $C^\infty$ function of $\{\bm{\eta}_m\}_{m \in \mathcal{M}_f}$ for every $\tau > 0$, and $\lim_{\tau \to \infty} \bm{\eta}_\tau^\star = \bm{\eta}_{m^\star}.$
}

\textit{Proof.} See Appendix~\ref{Appendix_Proof_Lemma_soft_theta}.\hfill$\blacksquare$

The reference $\bm{\eta}_\tau^\star$ is aligned with $\mathbf{k}$ and varies smoothly with the link geometry through the candidates~\eqref{eq:eta_candidates}; its magnitude is the soft-min-weighted distance to the nearest constructive-interference fringe. The temperature $\tau$ trades selectivity against smoothness: large $\tau$ recovers the hard $\arg\min$ at the cost of stiffer transitions, while small $\tau$ averages neighboring candidates and may slightly miss the optimal fringe. In Sec.~\ref{sec:numerical}, $\tau$ is tuned to the timescale of the geometric variations, and $\mathcal{M}$, $\mathcal{E}$ are sized so that $\mathcal{M}_f$ is nonempty throughout the mission. In the degenerate cases $\mathcal{M}_f = \varnothing$ or non-unique minimum-norm element, $\bm{\eta}_\tau^\star$ converges respectively to the projection of $\bm{\eta}_0$ onto $\mathcal{E}$ or to the centroid of the minimizing set, both continuous in the link geometry.



\section{Optimal Control Problem}
\label{sec:OCP}

The \ac{UAV} is modeled as a fully actuated \emph{Generalized Tilting Multirotor} (\acs{GTMR})~\cite{Ryll2019Interaction, Hamandi2021MRAV}, whose continuous-time rigid-body dynamics admit the standard Newton--Euler form $\dot{\mathbf{x}} = \mathbf{h}(\mathbf{x}, \mathbf{u})$, with state $\mathbf{x} = [\mathbf{p}^\top, \bm{\eta}^\top, \mathbf{v}^\top, \bm{\omega}^\top]^\top \in \mathbb{R}^{12}$ collecting position, Euler angles $\bm{\eta} = [\phi, \vartheta, \psi]^\top$, translational velocity, and body-frame angular velocity, and control input $\mathbf{u} = \bm{\Omega} \in \mathbb{R}^{N_p}$ given by the squared rotor speeds. Full actuation, ensured by the \acs{GTMR} geometry, decouples attitude regulation from translational authority, so the communication-driven attitude reference of Sec.~\ref{sec:Smooth_Control_Law} can be tracked without compromising the position trajectory.

The \ac{UAV} employs \ac{NMPC}~\cite{SilanoSMC2025, Silano2026CommNMPCJamming} to track the desired mission state subject to the dynamics, actuator limits, and operational constraints, embedding the communication-optimal attitude reference $\bm{\eta}_\tau^\star$ from~\eqref{eq:theta_soft_min} into the tracking objective at each control update.

\subsubsection{Discrete-time formulation and stage cost}
\label{subsec:OCP_cost}
Let $T_s \!>\! 0$ be the sampling period and $N$ the prediction horizon. The continuous-time dynamics are discretized via one-step integration $\mathbf{x}_{r+1} = \mathbf{h}_d(\mathbf{x}_r, \mathbf{u}_r)$ for $r \in \{0, \dots, N-1\}$, where $\mathbf{h}_d$ denotes the discrete-time state transition map. At each prediction step $r$, denote by $\mathbf{p}_{d,r}$ the desired position and by $\bm{\eta}_{d,r}^\star$ the communication-optimal attitude reference computed from~\eqref{eq:theta_soft_min} for the predicted geometry at step $r$.

The stage cost is the quadratic $\ell_r = \|\mathbf{p}_{d,r} \!- \mathbf{p}_r\|_{\mathbf{Q}_p}^2 + \|\bm{\eta}_{d,r}^\star - \bm{\eta}_r\|_{\mathbf{Q}_\eta}^2 + \|\mathbf{v}_r\|_{\mathbf{Q}_v}^2 + \|\bm{\omega}_r\|_{\mathbf{Q}_\omega}^2 + \|\mathbf{u}_r\|_{\mathbf{Q}_u}^2$, with $\mathbf{Q}_p, \mathbf{Q}_\eta \succ 0$ tracking weights and $\mathbf{Q}_v, \mathbf{Q}_\omega, \mathbf{Q}_u \succeq 0$ regularization weights that penalize aggressive maneuvers, smooth the actuator commands, and improve numerical conditioning. The terminal cost $\ell_N$ has the same form (without the input term) with weights $\mathbf{Q}_{p,N}, \mathbf{Q}_{\eta,N}, \mathbf{Q}_{v,N}, \mathbf{Q}_{\omega,N}$, yielding the finite-horizon objective $J = \sum_{r=0}^{N-1} \ell_r + \ell_N$.

\subsubsection{Optimal control problem}
\label{subsec:OCP_formulation}

Let $k \in \mathbb{N}$ index control updates at real time $t_k = kT_s$. At each $t_k$, the \ac{NMPC} solves
\vspace*{-0.75em}
\begin{subequations}\label{eq:opt_problem_OCP}
\small
\begin{align}
    \min_{\substack{\{\mathbf{u}_r\}_{r=0}^{N-1}}}\quad 
    & J = \sum_{r=0}^{N-1} \ell_r + \ell_N \\
    \text{s.t.}\quad 
    & \mathbf{x}_0 = \mathbf{x}(t_k), \\
    & \mathbf{x}_{r+1} = \mathbf{h}_d(\mathbf{x}_r, \mathbf{u}_r), 
    && \hspace{-0.5em}r = 0, \dots, N-1, \\
    & \mathbf{p}_r \in \mathcal{A},\;
      \bm{\eta}_r \in \mathcal{E}, 
    && \hspace{-0.5em}r = 0, \dots, N, \label{eq:OCP:state_box}\\
    & \underline{\mathbf{v}} \le \mathbf{v}_r \le \overline{\mathbf{v}}, \;
      \underline{\bm{\omega}} \le \bm{\omega}_r \le \overline{\bm{\omega}}, 
    && \hspace{-0.5em}r = 0, \dots, N, \label{eq:OCP:vel_box}\\
    & \underline{\bm{\Omega}} \le \mathbf{u}_r \le \overline{\bm{\Omega}}, 
    && \hspace{-0.5em}r = 0, \dots, N-1. \label{eq:OCP:input_box}
\end{align}
\normalsize
\end{subequations}
The set $\mathcal{A} \subset \mathbb{R}^3$ defines the admissible operational region; $\mathcal{E} \subset \mathbb{R}^3$ constrains the Euler angles to mechanically admissible values consistent with assumption~\textbf{(A1)}; and the bounds $\{\underline{\mathbf{v}}, \overline{\mathbf{v}}\}$, $\{\underline{\bm{\omega}}, \overline{\bm{\omega}}\}$, $\{\underline{\bm{\Omega}}, \overline{\bm{\Omega}}\}$ encode actuator and safety limits.

The first input $\mathbf{u}_0^\star$ of the optimal sequence is applied to the \ac{UAV}, and the problem is re-solved at $t_{k+1}$ in receding-horizon fashion. Real-time tractability is achieved through standard sequential quadratic programming with the real-time iteration scheme~\cite{Silano2026CommNMPCJamming, SilanoSMC2025}; the implementation details are given in Sec.~\ref{sec:numerical}.



\section{Numerical Results}
\label{sec:numerical}

We evaluate the proposed controller in a maritime \ac{A2S} scenario governed by the two-ray model. To exercise the 3D orientation law across a time-varying relative bearing, the \ac{UAV} tracks a horizontal circular trajectory of radius $R$ centered at $(x_0 - R, y_0, z_0)$, parameterized as $\mathbf{p}_d(t) = (x_0 - R + R\cos(2\pi s(t)),\, y_0 + R\sin(2\pi s(t)),\, z_0)$ with the smooth time-warp $s(t) = \tfrac{1}{2}\bigl[1 - \cos(\pi t/T)\bigr]$, $t \in [0, T]$, ensuring jerk-free start/stop kinematics. The ship moves at constant speed along the $x$-axis from $\mathbf{p}_{s,0}$ to $\mathbf{p}_{s,N}$. The communication-optimal reference $\bm{\eta}_{d,r}^\star$ is computed offline from~\eqref{eq:theta_soft_min} along the planned geometry and supplied to the \ac{NMPC} for online tracking.

Simulations run in MATLAB using the MATMPC toolbox\footnote{\url{https://github.com/chenyutao36/MATMPC}} with a fixed-step fourth-order Runge--Kutta integrator and qpOASES\footnote{\url{https://github.com/coin-or/qpOASES}} as the solver. Parameters are listed in Table~\ref{tab:SimParams}. We compare three controllers: the proposed 3D orientation-aware design, a pitch-only benchmark ($\phi = \psi = 0$, only $\vartheta$ optimized), and a zero-orientation baseline ($\bm{\eta} \equiv \mathbf{0}$). Performance is measured by the channel gain $|g(\bm{\eta})|^2$, the SNR-outage fraction across thresholds, and the cumulative throughput $C(t_k, \bm{\eta}) = \sum_{i=0}^{k} T_s \log_2\bigl(1 + P_t |g(\bm{\eta}(t_i))|^2/\sigma^2\bigr)$.

\begin{figure}[tb]
    \centering
    \begin{subfigure}{0.76\columnwidth}
        \centering
        \adjincludegraphics[
            width=\columnwidth,
            trim={{.0\width} {.06\height} {.0\width} {.11\height}},
            clip
        ]{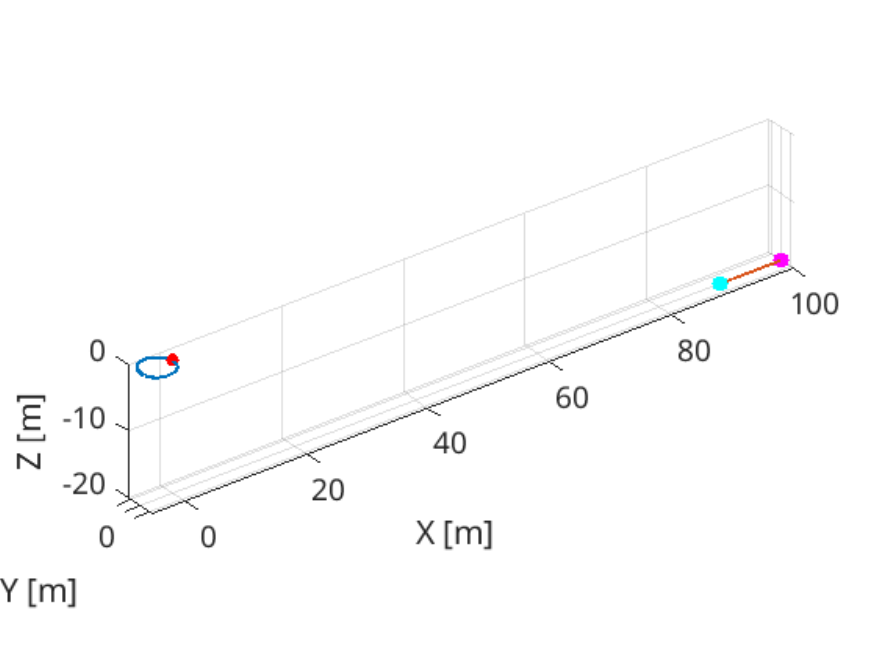}
        \vspace{-1.2em}
        \caption{3D trajectory of the \ac{UAV} and the ship.}
    \end{subfigure}
    \\
    \begin{subfigure}{0.76\columnwidth}
        \centering
        \adjincludegraphics[
            width=\columnwidth,
            trim={{.0\width} {.125\height} {.05\width} {.06\height}},
            clip
        ]{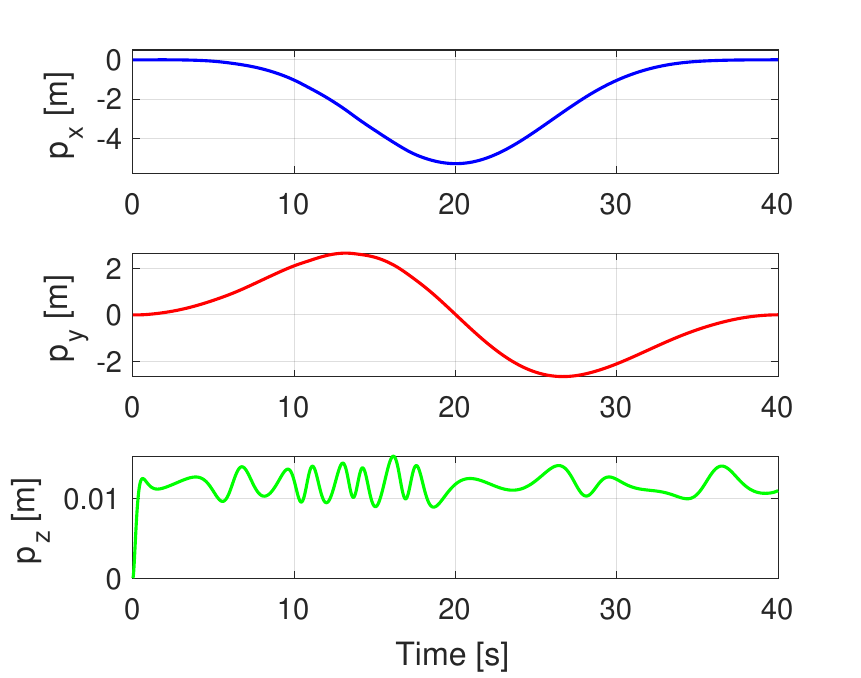}
        \vspace{-1.2em}
        \caption{Position coordinates versus time.}
    \end{subfigure}

    \vspace{-0.4em}
    \caption{3D and planar trajectories of the \ac{UAV} (blue) and the ship (red). Markers indicate start (green/cyan) and end (red/magenta) positions.}
    \label{fig:Traj_Pos}
    \vspace{-1.0em}
\end{figure}

\begin{figure}[t]
\vspace{4pt}
    \centering
    \begin{subfigure}{0.49\columnwidth}
        \centering
        \adjincludegraphics[
            width=\columnwidth,
            trim={{.0\width} {.015\height} {.05\width} {.05\height}},
            clip
        ]{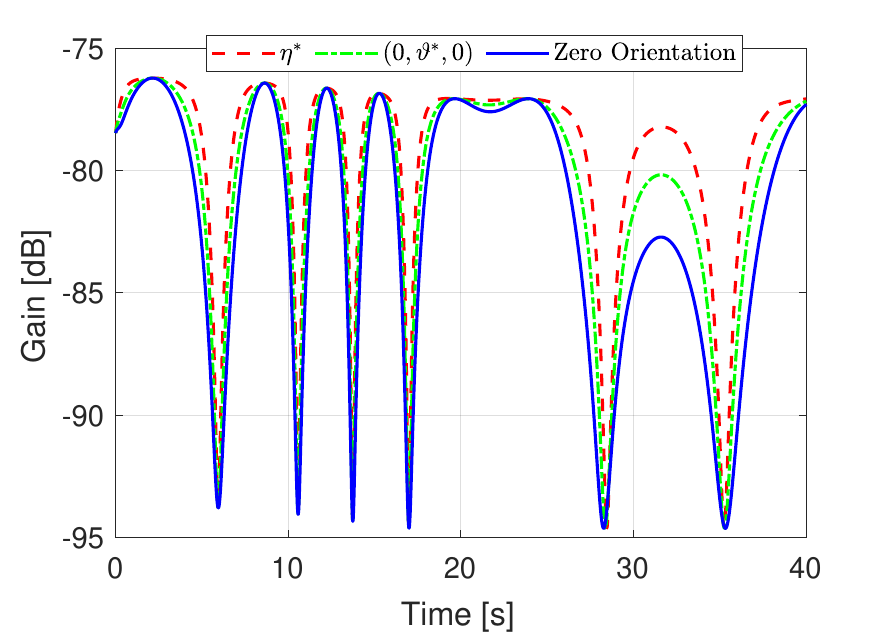}
        \vspace{-1.2em}
        \caption{}
        \label{fig:channel_gain}
    \end{subfigure}
    \hfill
    \begin{subfigure}{0.49\columnwidth}
        \centering
        \adjincludegraphics[
            width=\columnwidth,
            trim={{.0\width} {.015\height} {.05\width} {.00\height}},
            clip
        ]{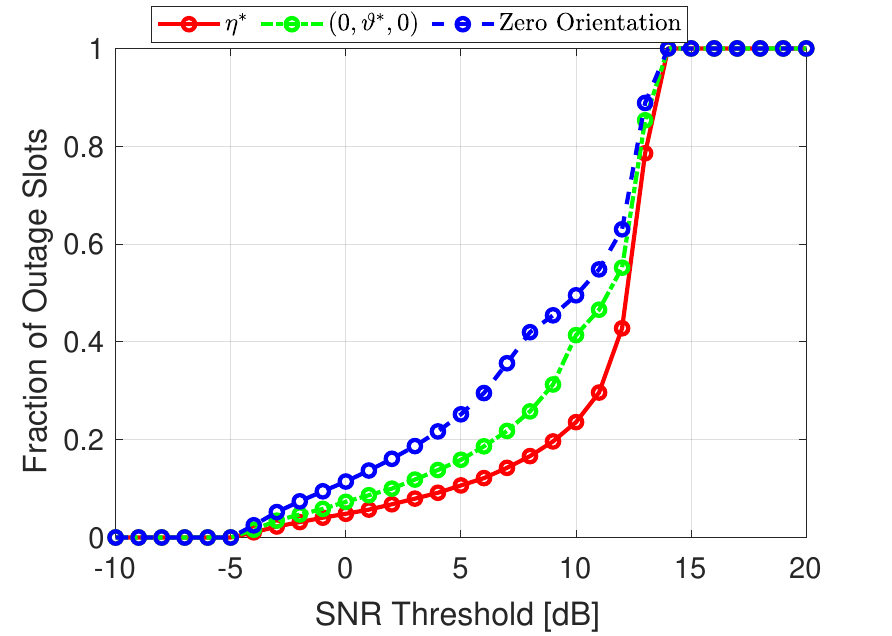}
        \vspace{-1.2em}
        \caption{}
        \label{fig:snr_outage}
    \end{subfigure}
    \vspace*{-1.5em}
    \caption{Channel gain (a) and SNR-outage fraction (b) for the three controllers.}
    \label{fig:Gains_and_Outage}
    \vspace{-1.0em}
\end{figure}

\textbf{Communication performance.}
Fig.~\ref{fig:Traj_Pos} confirms that all three controllers execute the prescribed circular trajectory while the ship moves laterally. As the relative bearing varies, the two-ray interference pattern sweeps through fringes, producing the deep, periodic fades visible in Fig.~\ref{fig:Gains_and_Outage}(a) for the zero-orientation baseline. The pitch-only benchmark partially mitigates these fades, whereas the 3D controller better aligns the antenna direction with $\mathbf{k}$ and sustains a higher channel gain over the mission. Using the channel-gain profiles reported in Fig.~\ref{fig:Gains_and_Outage}(a), we compute the total transmitted data over the whole mission. This calculation shows that the proposed scheme transmits $11.4\%$ and $22.2\%$ more data than the pitch-only and zero-orientation benchmarks, respectively. Fig.~\ref{fig:Gains_and_Outage}(b) reports the outage fraction over a range of \ac{SNR} thresholds: the 3D controller dominates uniformly, indicating not only higher mean gain but also fewer deep-fade events that cause loss-of-link.

\begin{table}[tb]
\vspace{4pt}
    \centering
    \caption{Simulation parameters}
    \vspace{-0.6em}
    \label{tab:SimParams}
    \renewcommand{\arraystretch}{1.05}
    \begin{adjustbox}{max width=0.98\columnwidth}
    \begin{tabular}{l c}
    \hline
    \multicolumn{2}{c}{\textbf{\ac{UAV} platform}} \\
    \hline
    Mass $m$ & $2.57\ \mathrm{kg}$ \\
    Inertia $\mathbf{J}$ & $\mathrm{diag}(0.11,\,0.11,\,0.19)\ \mathrm{kg\cdot m}^2$ \\
    Antenna offset $\mathbf{r}_a$ & $\left[0.39,\, 0.39,\, 0.195\right]^\top\ \mathrm{m}$ \\
    Rotor speed bounds $[\underline{\bm{\Omega}},\,\bar{\bm{\Omega}}]$ & $[16, 100]\ \mathrm{Hz}$ \\
    \hline
    \multicolumn{2}{c}{\textbf{Channel and scenario}} \\
    \hline
    Carrier $f_c$, reflection $\gamma$ & $3\ \mathrm{GHz}$, $0.8$ \\ 
    Ship antenna height $h$ & $10\ \mathrm{m}$\\
    Ship endpoints $\mathbf{p}_{s,0}$, $\mathbf{p}_{s,N}$ & $\left[90,\,0,\,-20\right]^\top,\ \left[100,\,0,\,-20\right]^\top\ \mathrm{m}$\\
    Trajectory radius $R$, duration $T$ & $2.65\ \mathrm{m}$, $40\ \mathrm{s}$ \\
    Transmit power $P_t$, noise $\sigma^2$ & $10\ \mathrm{dBm}$, $-80\ \mathrm{dBm}$ \\
    \hline
    \multicolumn{2}{c}{\textbf{\ac{NMPC}}} \\
    \hline
    Sampling $T_s$, horizon $N$ & $0.02\ \mathrm{s}$, $30$ \\
    Position weight $\mathbf{Q}_p\!=\!\mathbf{Q}_{p,N}$ & $100\,\mathbf{I}_3$ \\
    Attitude weight $\mathbf{Q}_\eta\!=\!\mathbf{Q}_{\eta,N}$ & $50\,\mathbf{I}_3$ \\
    Velocity weight $\mathbf{Q}_v\!=\!\mathbf{Q}_{v,N}$ & $\mathbf{I}_3$ \\
    Angular-rate weight $\mathbf{Q}_\omega\!=\!\mathbf{Q}_{\omega,N}$ & $10\,\mathbf{I}_3$ \\
    Input weight $\mathbf{Q}_u$ & $3.2\!\times\!10^{-8}\,\mathbf{I}_6$ \\
    Velocity bounds $[\underline{\mathbf{v}},\,\overline{\mathbf{v}}]$ & $\pm[1,\,1,\,1]^\top\ \mathrm{m/s}$ \\
    Angular-rate bounds $[\underline{\bm{\omega}},\,\overline{\bm{\omega}}]$ & $\pm[1,\,1,\,1]^\top\ \mathrm{rad/s}$ \\
    Euler-angle bounds $\mathcal{E}$ & $\pm[15,\,15,\,15]^\top\ \mathrm{deg}$ \\ 
    \hline
    \multicolumn{2}{c}{\textbf{Soft-min hyperparameters}} \\
    \hline
    Temperature $\tau$, candidate set $\mathcal{M}$ & $7$, $\{-50,\,\ldots,\,50\}$ \\
    \hline
    \end{tabular}
    \end{adjustbox}
    \vspace*{-1.5em}
\end{table}


\begin{figure}[tb]
\vspace{4pt}
    \centering
    \begin{subfigure}{0.70\columnwidth}
    \centering
        \IfFileExists{figures/Results/Globecom_V1/Optimal_Euler_Angles_Vs_Time_Plot_MobUAV_Scenario.pdf}{%
        \adjincludegraphics[width=\columnwidth, trim={{.0\width} {.015\height} {.05\width} {.05\height}}, clip]{figures/Results/ICUAS_V1/Optimal_Euler_Angles_Vs_Time_Plot_MobUAV_Scenario.pdf}%
        }{%
        \adjincludegraphics[width=\columnwidth, trim={{.0\width} {.015\height} {.05\width} {.05\height}}, clip]{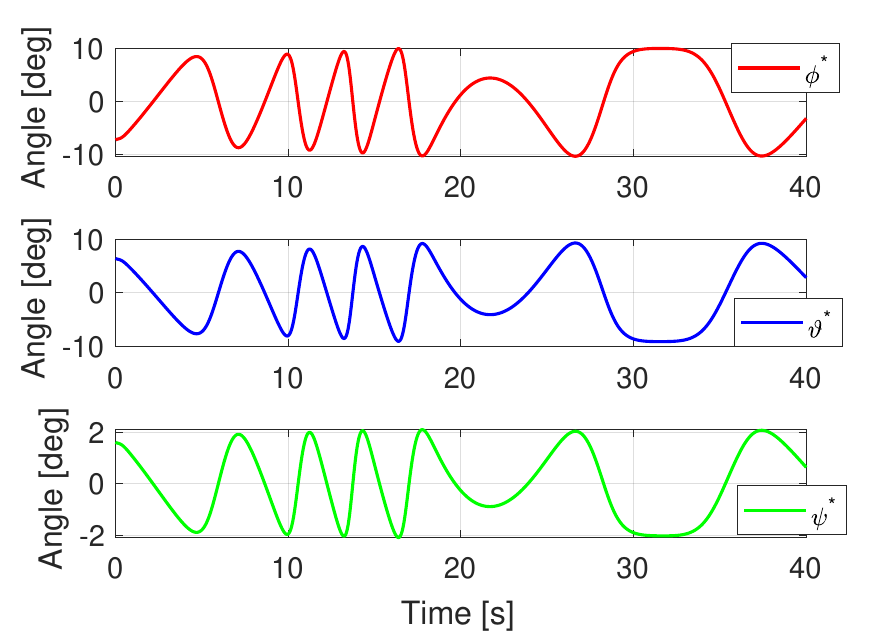}%
        }
    \vspace{-1.5em}
    \caption{Communication-optimal Euler-angles reference.}
    \end{subfigure}
    \\
    \begin{subfigure}{0.70\columnwidth}
    \centering
        \IfFileExists{figures/Results/ICUAS_V1/Tracking_Error_Euler_Angles_MobUAV_Scenario.pdf}{%
        \adjincludegraphics[width=\columnwidth, trim={{.0\width} {.015\height} {.05\width} {.05\height}}, clip]{figures/Results/Globecom_V1/Tracking_Error_Euler_Angles_MobUAV_Scenario.pdf}%
        }{%
        \adjincludegraphics[width=\columnwidth, trim={{.0\width} {.015\height} {.04\width} {.05\height}}, clip]{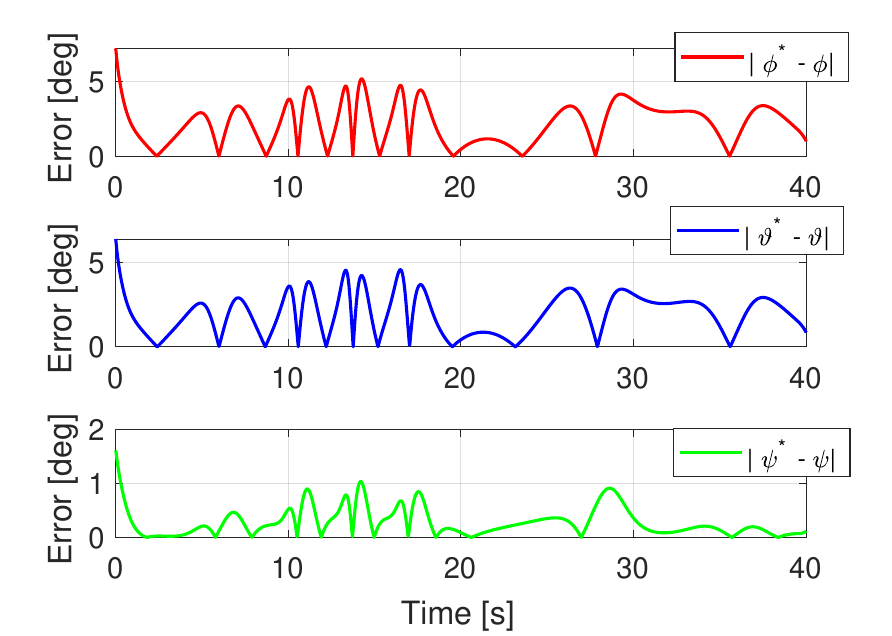}%
        }
    \vspace{-1.5em}
    \caption{Euler-angles tracking errors.}
    \end{subfigure}
    \vspace{-0.25em}
    \caption{Attitude tracking performance of the 3D communications-aware orientation law.}
    \label{fig:Angles_OpitmalThet_Error}
    \vspace{-1em}
\end{figure}

\begin{figure}[tb]
    \centering
    \adjincludegraphics[width=0.70\columnwidth, trim={{.0\width} {.0\height} {.05\width} {.05\height}}, clip]{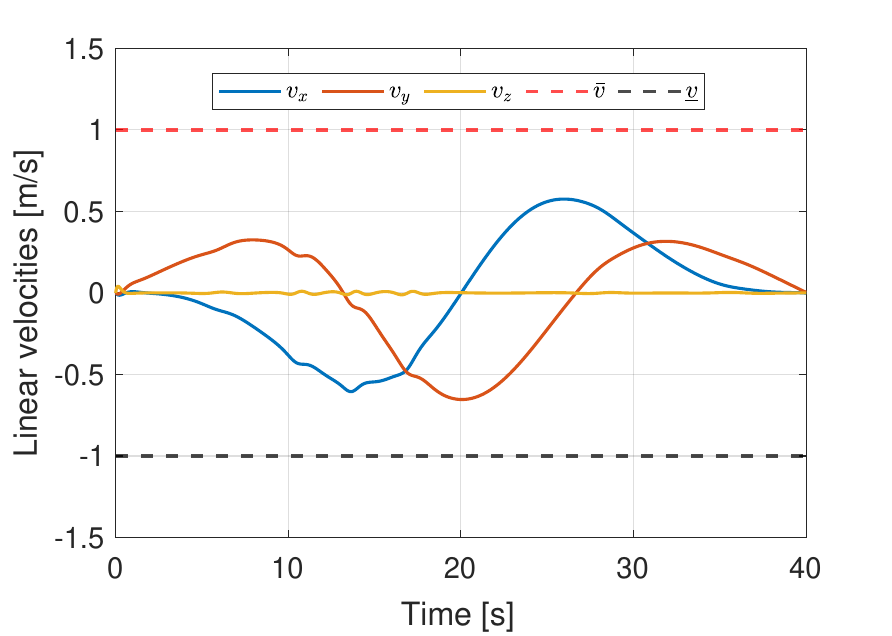}
    \vspace{-0.5em}
    \caption{Linear velocity components.}
    \label{fig:Velocities}
    \vspace{-1em}
\end{figure}

\begin{figure}[tb]
    \centering
    \adjincludegraphics[width=0.70\columnwidth, trim={{.0\width} {.0\height} {.05\width} {.05\height}}, clip]{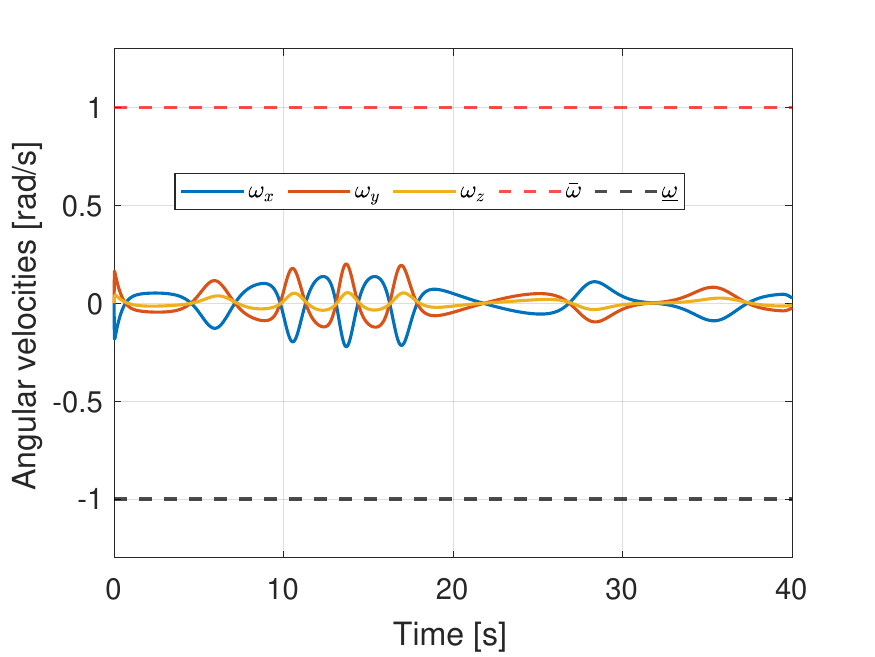}
    \vspace{-0.5em}
    \caption{Angular velocity components.}
    \label{fig:Ang_Vel_MobUAV}
    \vspace{-1em}
\end{figure}

\begin{figure}[tb]
    \centering
    \adjincludegraphics[width=\columnwidth, trim={{.0\width} {.025\height} {.0\width} {.00\height}}, clip]{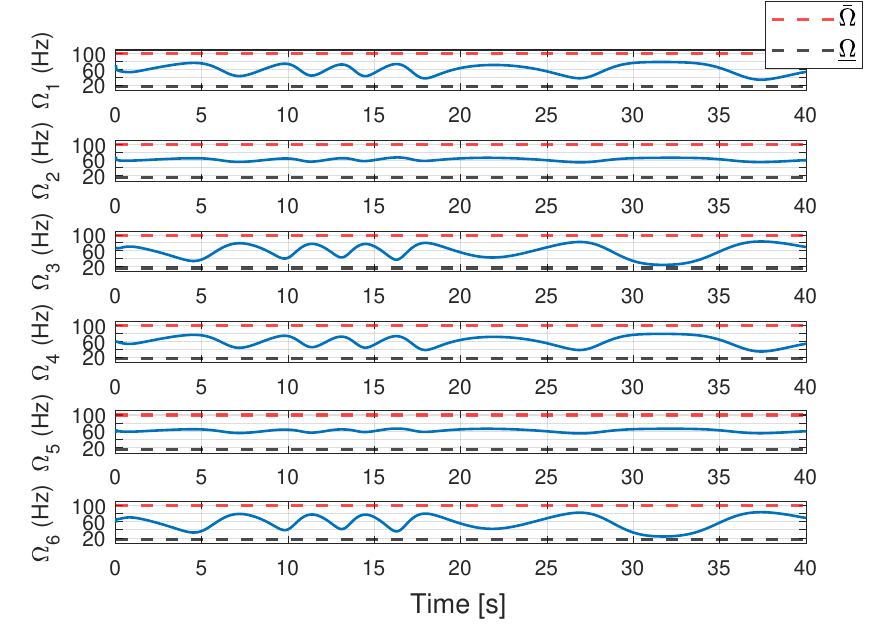}
    \vspace*{-1.1em}
    \caption{Rotor speed commands.}
    \label{fig:RotorSpeeds_MobUAV}
\end{figure}

\textbf{Attitude tracking and feasibility.}
Fig.~\ref{fig:Angles_OpitmalThet_Error} compares the tracked Euler angle errors against the offline reference. A non-negligible tracking residual is visible: the reference fluctuates on a timescale faster than the closed-loop bandwidth permits under the actuator and small-angle constraints. This is the expected trade-off between tracking fidelity and dynamical feasibility---and notably, partial tracking already realizes the bulk of the achievable communication gain, since the controller still steers the antenna phase center along $\mathbf{k}$, just with a delay. Figs.~\ref{fig:Velocities}--\ref{fig:RotorSpeeds_MobUAV} confirm feasibility throughout the mission: linear velocities, angular rates, and rotor speeds remain within their bounds, and actuator profiles evolve smoothly without chattering.

The 3D formulation generalizes the vertical-plane model, exploits the full attitude authority of the \acs{GTMR}, and improves real-time link quality without sacrificing the primary motion task.



\section{Conclusions}
\label{sec:conclusions}

This paper treated \ac{UAV} attitude as a control variable for shaping the two-ray interference in maritime \ac{A2S} links. Under small-angle and far-field assumptions, the constructive-interference condition reduces to an affine constraint in the Euler angles, yielding closed-form minimum-norm attitude candidates governed by a geometry-dependent phase-sensitivity vector. A differentiable soft-minimum rule converts this discrete family into a smooth reference tracked by a constrained \ac{NMPC} controller on a fully-actuated tilting multirotor. Simulations of a maritime scenario showed cumulative throughput gains of $11.4\%$ and $22.2\%$ over pitch-only and zero-orientation benchmarks, while preserving trajectory tracking and respecting actuator limits. Future work will extend the model to rough sea states, time-varying reflection coefficients, pose-estimation uncertainty, and onboard experimental validation.



\appendix

\subsection{Proof of Theorem 1}
\label{Appendix_Proof_Theorem}

\textit{Affine phase model.} 
From~\eqref{eq:antenna_phase_center_3d}--\eqref{eq:small_angle_rotation}, $\mathbf{p}_a(\bm{\eta}) - \mathbf{p}_{a,0} = \mathbf{J}_a\bm{\eta} + \mathcal{O}(\|\bm{\eta}\|^2)$. For any fixed $\mathbf{q} \neq \mathbf{p}_{a,0}$, the distance $d_\mathbf{q}(\bm{\eta}) = \|\mathbf{q} - \mathbf{p}_a(\bm{\eta})\|$ admits the first-order expansion $d_\mathbf{q}(\bm{\eta}) = d_\mathbf{q}(\mathbf{0}) - \mathbf{u}_\mathbf{q}^\top \mathbf{J}_a\bm{\eta} + \mathcal{O}(\|\bm{\eta}\|^2)$, where $\mathbf{u}_\mathbf{q} = (\mathbf{q} - \mathbf{p}_{a,0})/\|\mathbf{q} - \mathbf{p}_{a,0}\|$. Specializing to $\mathbf{q} \in \{\mathbf{p}_s, \mathbf{p}_s^{\rm im}\}$ and substituting into~\eqref{eq:DeltaPhi_eta} yields
\begin{equation}
    \Delta\Phi(\bm{\eta}) = \Delta\Phi_0 + \tfrac{2\pi}{\lambda}(\mathbf{u}_{\rm{LoS}} - \mathbf{u}_{\rm{ref}})^\top \mathbf{J}_a\bm{\eta} + \mathcal{O}(\|\bm{\eta}\|^2),
\end{equation}
which is~\eqref{eq:phase_affine} by the definition of $\mathbf{k}$ in~\eqref{eq:k_vector}.

\textit{Minimum-norm constructive-interference attitudes.}
The coefficient of $\cos(\Delta\Phi(\bm{\eta}))$ in~\eqref{eq:power_expand} is negative, so maximizing $|g(\bm{\eta})|^2$ at leading order is equivalent to enforcing $\Delta\Phi(\bm{\eta}) = (2m+1)\pi$, equivalently $\mathbf{k}^\top\bm{\eta} = C_m$ via~\eqref{eq:phase_affine}. The minimum-norm element on this hyperplane solves $\min_{\bm{\eta} \in \mathbb{R}^3} \tfrac{1}{2}\|\bm{\eta}\|^2$ s.t.\ $\mathbf{k}^\top\bm{\eta} = C_m$, a strictly convex QP with KKT solution $\bm{\eta}_m = (C_m/\|\mathbf{k}\|^2)\mathbf{k}$, which is~\eqref{eq:eta_candidates}. \hfill$\blacksquare$

\subsection{Proof of Lemma 1}
\label{Appendix_Proof_Lemma_soft_theta}

The soft weights $w_m(\tau) = e^{-\tau\|\bm{\eta}_m\|^2}/\sum_{\ell \in \mathcal{M}_f} e^{-\tau\|\bm{\eta}_\ell\|^2}$ satisfy $w_m(\tau) > 0$ and $\sum_m w_m(\tau) = 1$, so~\eqref{eq:theta_soft_min} is the convex combination $\bm{\eta}_\tau^\star = \sum_{m \in \mathcal{M}_f} w_m(\tau)\,\bm{\eta}_m$, smooth in $\{\bm{\eta}_m\}$ for every $\tau > 0$. Let $m^\star$ be the unique minimizer of $\|\bm{\eta}_m\|$. For $m \neq m^\star$, $w_m(\tau)/w_{m^\star}(\tau) = \exp(-\tau(\|\bm{\eta}_m\|^2 - \|\bm{\eta}_{m^\star}\|^2)) \to 0$ as $\tau \to \infty$, so $w_{m^\star}(\tau) \to 1$ and $\bm{\eta}_\tau^\star \to \bm{\eta}_{m^\star}$. \hfill$\blacksquare$



\bibliographystyle{IEEEtran}
\bibliography{Biblio_UAV_TwoRay.bib}

\end{document}